\documentclass[sigconf]{acmart}

\usepackage[utf8]{inputenc} 
\usepackage[T1]{fontenc}    
\usepackage{hyperref}       
\usepackage{url}            
\usepackage{booktabs}       
\usepackage{amsfonts}       
\usepackage{nicefrac}       
\usepackage{microtype}      
\usepackage{xcolor}         

\usepackage{threeparttable}
\usepackage{multirow}
\usepackage{graphicx}
\usepackage{newtxmath}
\usepackage{rotating}

\usepackage{wrapfig}
\usepackage{enumitem}
\usepackage{ragged2e}
\usepackage{subcaption}

\usepackage{CJKutf8}         

\usepackage{fontawesome5} 
\usepackage{academicons}  
\usepackage{xcolor}       

\usepackage{threeparttable}
\usepackage{multirow}
\usepackage{graphicx}
\usepackage{newtxmath}
\usepackage{rotating}

\usepackage{wrapfig}
\usepackage{enumitem}
\usepackage{ragged2e}
\usepackage{listings}

\usepackage{arydshln}
\usepackage{makecell}
\usepackage[table]{xcolor}

\usepackage[capitalize,noabbrev]{cleveref}
\usepackage{lipsum}
\usepackage{booktabs}
\usepackage{multirow}
\usepackage{array}
\usepackage{arydshln}
\usepackage{colortbl}
\usepackage{fontawesome5}
\usepackage[utf8]{inputenc}
\usepackage{tabularx}   
\usepackage{amsmath}    
\usepackage{pifont}      
\usepackage{algorithm}
\usepackage{algpseudocode} 
\usepackage{booktabs}
\usepackage{xcolor}
\usepackage{makecell}

\definecolor{tabblue}{HTML}{1f77b4}
\definecolor{taborange}{HTML}{ff7f0e}
\definecolor{tabgreen}{HTML}{2ca02c}
\definecolor{tabred}{HTML}{d62728}
\definecolor{tabpurple}{HTML}{9467bd}
\definecolor{tabpink}{HTML}{ff0080}

\newcolumntype{C}{>{\centering\arraybackslash}X}

\newcommand{\up}[1]{\textcolor{tabpink}{\scriptsize$\uparrow$#1}}
\newcommand{\down}[1]{\textcolor{tabgreen}{\scriptsize$\downarrow$#1}}

\newcommand{\scoreup}[2]{\shortstack[c]{\rule{0pt}{2.5ex}#1\\[-1pt]\up{#2}}}
\newcommand{\scoredown}[2]{\shortstack[c]{\rule{0pt}{2.5ex}#1\\[-1pt]\down{#2}}}

\AtBeginDocument{%
  }

\copyrightyear{2026}
\acmYear{2026}
\setcopyright{cc}
\setcctype{by}
\acmConference[MM '26]{Proceedings of the 34th ACM International Conference on Multimedia}{November 10--14, 2026}{Rio de Janeiro, Brazil}
\acmBooktitle{Proceedings of the 34th ACM International Conference on Multimedia (MM '26), November 10--14, 2026, Rio de Janeiro, Brazil}
\acmDOI{10.1145/3767308.3835620}
\acmISBN{979-8-4007-2213-4/2026/11}

\begin{document}

\title{Reward-Aware Trajectory Shaping for Few-step Visual Generation}

\author{Rui Li}
\authornote{Both authors contributed equally to this research.}
\email{rui.li@mail.ustc.edu.cn}
\authornotemark[1]
\affiliation{%
  \institution{University of Science and Technology of China}
  \city{HeFei}
  \country{China}
}
\affiliation{%
  \institution{Institute of Artificial Intelligence, China Telecom (TeleAI)}
  \city{Shanghai}
  \country{China}
}
\author{Bingyu Li}
\authornote{Both authors contributed equally to this research.}
\email{libingyu0205@mail.ustc.edu.cn}
\authornotemark[1]
\affiliation{%
  \institution{University of Science and Technology of China}
  \city{HeFei}
  \country{China}
}
\affiliation{%
  \institution{Institute of Artificial Intelligence, China Telecom (TeleAI)}
  \city{Shanghai}
  \country{China}
}

\author{Yuanzhi Liang}
\affiliation{%
  \institution{Institute of Artificial Intelligence, China Telecom (TeleAI)}
  \city{Shanghai}
  \country{China}
}

\author{Haibin Huang}
\affiliation{%
  \institution{Institute of Artificial Intelligence, China Telecom (TeleAI)}
  \city{Shanghai}
  \country{China}
}

\author{Chi Zhang}
\affiliation{%
  \institution{Institute of Artificial Intelligence, China Telecom (TeleAI)}
  \city{Shanghai}
  \country{China}
}
\author{XueLong Li}
\authornote{Corresponding author.}
\email{xuelong\_li@ieee.org
}
\affiliation{%
  \institution{Institute of Artificial Intelligence, China Telecom (TeleAI)}
  \city{Shanghai}
  \country{China}
}

\renewcommand{\shortauthors}{Rui Li et al.}

\begin{abstract}
Achieving high-fidelity generation in extremely few sampling steps has long been a central goal of generative modeling. Existing approaches largely rely on distillation-based frameworks to compress the original multi-step denoising process into a few-step generator. However, such methods inherently constrain the student to imitate a stronger multi-step teacher, imposing the teacher as an upper bound on student performance. We argue that introducing \textbf{preference alignment awareness} enables the student to optimize toward reward-preferred generation quality, potentially surpassing the teacher instead of being restricted to rigid teacher imitation.
To this end, we propose \textbf{Reward-Aware Trajectory Shaping (RATS)}, a lightweight framework for preference-aligned few-step generation without the need of additional data. Specifically, teacher and student latent trajectories are aligned at key denoising stages through horizon matching, while a \textbf{reward-aware gate} is introduced to adaptively regulate teacher guidance based on their relative reward performance. Trajectory shaping is strengthened when the teacher achieves higher rewards, and relaxed when the student matches or surpasses the teacher, thereby enabling continued reward-driven improvement. By seamlessly integrating trajectory distillation, reward-aware gating, and preference alignment, RATS effectively transfers preference-relevant knowledge from high-step generators without incurring additional test-time computational overhead. Experimental results demonstrate that RATS exhibits strong few‑step generation capability for visual generation while achieving a favorable efficiency–quality trade‑off.
\end{abstract}




%
\maketitle

\section{Introduction}
\begin{figure}[t]
    \centering
    \includegraphics[width=0.95\linewidth]{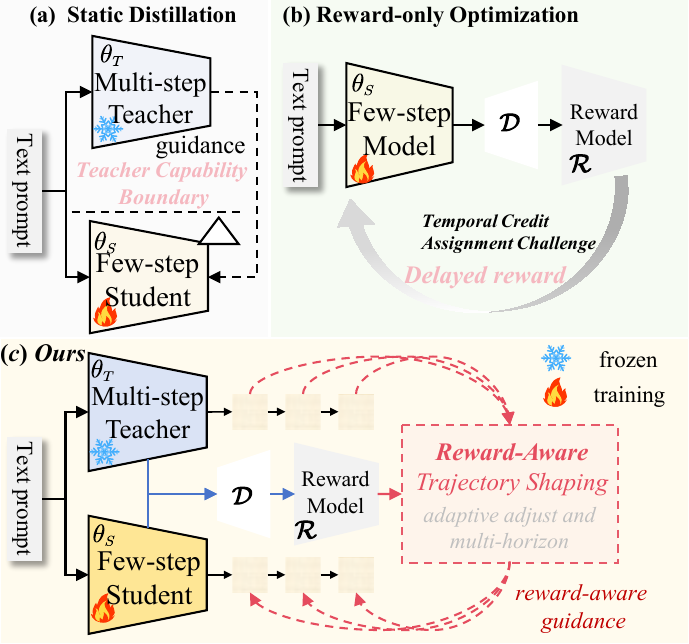}
    \caption{Comparison of our framework with existing few-step generation and reward optimization paradigms.}
    \label{fig:pic_intro}
    \Description{1}
\end{figure}

Generative models have achieved remarkable progress in visual synthesis, producing highly realistic and diverse outputs across both image and video domains~\cite{kong2024hunyuanvideo,shi2024motion,wan2025wan}. However, despite these advances, high-quality synthesis still typically relies on iterative denoising with many sampling steps, resulting in substantial inference cost. This issue is particularly severe in visual generation, where the high dimensionality of spatiotemporal outputs makes long denoising trajectories computationally prohibitive~\cite{kong2024hunyuanvideo,wan2025wan,ren2024hyper,ge2025senseflow}. Consequently, enabling high-quality generation under an extremely limited step budget has become a central problem in efficient visual synthesis~\cite{an2026ai,liang2026integrating,zhang2024vast,liang2026teleboost,liu2026learning,li2025rethinking,liang2022seeg,liang2022simple,li2022positive,zhang2025variational,chen2026generative}.

Recent progress in few-step generation has shown that long denoising trajectories can be aggressively compressed while retaining competitive visual quality~\cite{salimans2022progressive,luhman2021knowledge,luo2023latent,song2023consistency,yin2024one,yin2024improved,sauer2024adversarial,EMdistillation,lin2024sdxl,chen2025sana,li2024t2v,luo2025enhance}. However, most existing methods remain fundamentally distillation-driven: a few-step student is trained to imitate a stronger many-step generator under a fixed step budget. This makes the student highly effective at trajectory compression, yet largely bounded by the capability of the teacher. In other words, the main limitation is no longer efficiency alone, but the fact that few-step students are typically optimized to imitate, rather than to align (shown in Fig.~\ref{fig:pic_intro}(a)).

A natural way to overcome this teacher-imposed ceiling is to directly optimize the few-step student toward human preferences with reward-based post-training. Such methods have recently shown strong ability to improve text faithfulness, aesthetic quality, and overall preference alignment in visual generation~\cite{ddpo,diffusiondpo,prabhudesai2024video,yuan2024instructvideo,liu2025improving,yang2025ipo,liu2025flow,dancegrpo}. However, directly applying terminal reward optimization to the few-step regime introduces a fundamental difficulty (shown in Fig.~\ref{fig:pic_intro}(b)): although reward signals are defined on final outputs, from the perspective of the denoising process they are only observed after the full rollout and therefore act as delayed supervision over the trajectory. Prior work has shown that diffusion alignment with reinforcement learning can be significantly affected by sparse or delayed rewards, leading to temporal credit assignment difficulties~\cite{ranzato2015sequence,schultz1997neural,doya2000reinforcement,karwowski2023goodhart,he2025tempflow,wang2026tagrpo}. This challenge becomes especially severe under extreme step compression, where each denoising step must simultaneously support both synthesis and alignment\cite{xu2025stare, shen2025fine, li2026toward, shen2026egoforge}.

These observations suggest that few-step alignment should not be viewed merely as an output-level reward optimization problem, but as a \emph{trajectory-level reward allocation} problem. Therefore, the central question is not only how to optimize a few-step generator with rewards, but also how to provide informative intermediate guidance so that preference signals can be effectively propagated across an aggressively compressed denoising trajectory.

Motivated by the comparison in Fig.~\ref{fig:pic_intro}, we propose \textbf{Reward-Aware Trajectory Shaping (RATS)}, a training framework for preference-aligned few-step generation. Our core insight is that intermediate denoising knowledge from a stronger many-step generator should be transferred to the few-step student only when it remains beneficial under the current reward objective, thereby allowing the student to benefit from teacher guidance without being constrained by it. Concretely, our method employs a few-step student~\cite{wan2025wan} as the sole generator at inference time, while introducing a multi-step exponential moving average (EMA) teacher only during training to provide multi-stage latent guidance. We adopt a sigma-based horizon matching strategy to align teacher--student latent trajectories across multiple denoising stages. More importantly, we introduce a \emph{reward gate} that dynamically modulates the strength of teacher guidance according to relative reward performance: when the teacher achieves a higher reward, trajectory shaping is strengthened to provide informative intermediate priors; when the student approaches or even surpasses the teacher, the teacher constraint is automatically weakened. In this way, the teacher acts as a conditional source of preference-relevant trajectory knowledge rather than a rigid target for imitation.

As a result, our framework substantially improves the efficiency--quality frontier in few-step generation. It significantly narrows the gap between a few-step student and stronger multi-step generators, while fully preserving the deployment efficiency of the student model. In summary, this work makes the following principal contributions:
\begin{itemize}
\item We systematically investigate the \textbf{teacher-bounded} bottleneck in preference-aligned few-step generation, and reformulate the problem as \textbf{reward-aware trajectory shaping}.
\item We introduce \textbf{RATS}, a training paradigm that conditionally transfers intermediate knowledge from a multi-step EMA teacher through a \textbf{reward gate}, enabling the student to go beyond static teacher imitation.
\item We instantiate RATS as a lightweight unified few-step visual generation framework without the need of additional data.
\item Extensive experiments demonstrate that \textbf{RATS} significantly improves the efficiency--quality trade-off, consistently outperforming existing baselines and substantially narrowing the gap to multi-step generators.
\end{itemize}

\section{Related Work}

\subsection{Few-Step Visual Generation}
Improving the efficiency of diffusion and flow-based generative models has become a central topic in visual generation. Existing approaches, including improved samplers, progressive or adversarial distillation, consistency-based modeling, and post-training acceleration strategies, have shown that long denoising trajectories can be compressed into only a few sampling steps while preserving competitive generation quality~\cite{salimans2022progressive,luhman2021knowledge,luo2023latent,song2023consistency,kimctm,yin2024one,yin2024improved,sauer2024adversarial,EMdistillation,chadebec2025flash,ren2024hyper,lin2024sdxl,yu2025visualizing, yu2026dinov3, yu2026spatiotemporal,lu2024simplifying,chen2025sana,shao2025rayflow,li2024t2v,luo2025enhance}. In video generation, few-step acceleration is especially important because the high dimensionality of spatiotemporal outputs makes multi-step sampling particularly expensive~\cite{kong2024hunyuanvideo,wan2025wan}.
Despite their effectiveness, existing few-step methods are primarily designed for trajectory compression rather than downstream alignment. Their objective is typically to reproduce the behavior of a stronger many-step generator as faithfully as possible under a constrained sampling budget. Consequently, while they improve efficiency, they offer limited flexibility for the few-step model to deviate from or improve beyond the teacher when optimization is driven by preference-aligned objectives. In contrast, our work studies few-step generation from the perspective of preference alignment, and explicitly addresses the limitation of teacher-bounded distillation.

\subsection{Visual Preference Alignment}
Reward-based post-training has recently emerged as an effective paradigm for improving text faithfulness, aesthetic quality, and human preference alignment in diffusion and flow-based visual generation~\cite{ddpo,diffusiondpo,prabhudesai2024video,yuan2024instructvideo,liu2025improving,yang2025ipo,liu2025flow,dancegrpo}. These methods typically optimize generation quality using external reward signals, often instantiated through learned preference or reward models~\cite{christiano2017deep,kirstain2023pick,xu2023imagereward,wu2023human,xu2024visionreward,zhang2025r1}. More recent GRPO-style methods further demonstrate the promise of online reward optimization for visual generation~\cite{liu2025flow,he2025tempflow,li2025mixgrpo,wang2025pref,wang2026tagrpo,dancegrpo}.
Taken together, these advances reveal a gap between two research directions: few-step generation focuses on efficient teacher imitation, while visual preference alignment focuses on reward-driven output optimization. Our work lies at their intersection, where the central challenge is how to align a heavily compressed denoising trajectory without remaining bounded by the teacher.
However, directly applying reward optimization to few-step generation remains challenging. Although reward signals can supervise final outputs, they are only defined after the full rollout and therefore act as delayed supervision over the denoising trajectory. Prior work has shown that sparse or delayed rewards can lead to temporal credit assignment difficulties in sequential optimization~\cite{ranzato2015sequence,schultz1997neural,doya2000reinforcement,karwowski2023goodhart}. This issue becomes even more severe in the few-step regime, where each denoising step carries a larger share of both synthesis and alignment. Our work addresses this problem by complementing terminal reward optimization with intermediate trajectory guidance from a stronger many-step generator, enabling preference-aware shaping of the few-step denoising process.

\section{Method}
\label{sec:method}

\begin{figure*}[ht]
    \centering
    \includegraphics[width=0.95\linewidth]{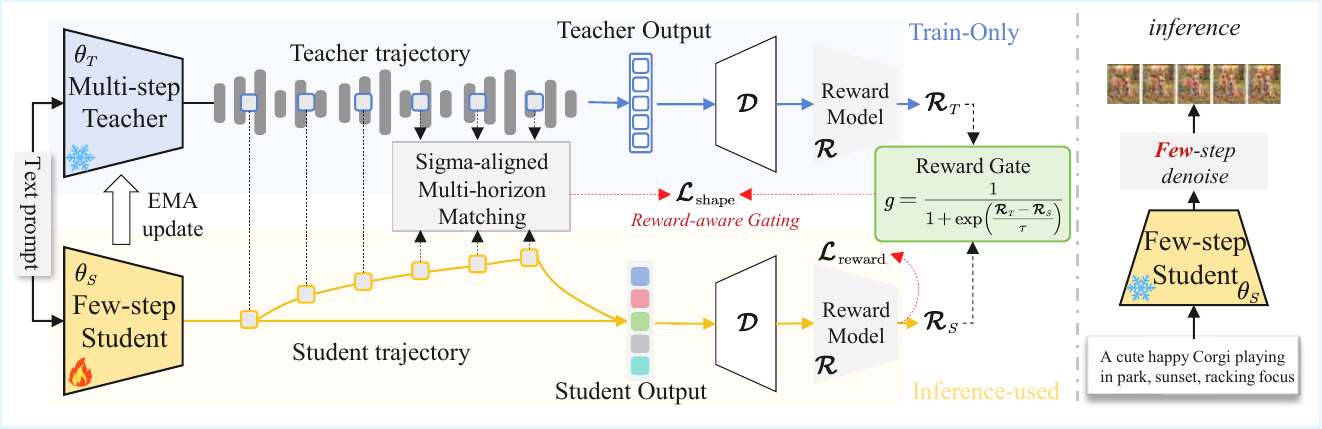}
    \caption{\textbf{Overview of Reward-Aware Trajectory Shaping (RATS).}
Given the same initial noise and text prompt, a few-step student is rolled out during training. At multiple student noise levels, the multi-step EMA teacher completes the corresponding student states toward clean endpoints, yielding sigma-aligned, state-conditioned targets. These targets jointly shape the student trajectory in latent and decoded image spaces, together producing $\mathcal{L}_{\text{shape}}$. The final outputs are further evaluated by a reward model, and a reward gate dynamically controls the complete shaping objective based on the relative rewards of teacher and student. During inference, only the few-step student is used, introducing no extra cost.}
    \label{fig:pic}
    \Description{2}
\end{figure*}

We present \textbf{Reward-Aware Trajectory Shaping (RATS)} shown in Figure \ref{fig:pic}, a training framework that reframes few-step generation from static teacher imitation to \emph{reward-aware trajectory shaping}.
Rather than training a few-step student to unconditionally reproduce a many-step teacher, RATS selectively transfers intermediate denoising knowledge only when it remains beneficial under the current reward objective, enabling the student to inherit useful high-step guidance while retaining the freedom to surpass the teacher.
A multi-step EMA teacher and the auxiliary shaping signals are used \emph{exclusively during training}; at inference time, only the lightweight few-step student with its native sampler is deployed, incurring zero additional cost.
We realize this idea through three interlocking mechanisms:
\textbf{(i)} an \emph{EMA teacher} that adapts with the student, providing non-stationary yet stable trajectory references;
\textbf{(ii)} \emph{sigma-aligned multi-horizon shaping} that constructs state-conditioned teacher endpoints at selected student noise levels and jointly shapes the student in latent and decoded image spaces;
and \textbf{(iii)} a \emph{reward gate} that dynamically modulates the complete shaping objective based on relative reward performance, enabling the student to surpass the teacher when it is ready.
All components are evaluated from the same student rollout and its state-conditioned teacher completions, and are optimized jointly within each training iteration.
These components are detailed in \cref{sec:shaping,sec:gate,sec:objective}.

\noindent\textbf{Notation.}
Let $\{\mathbf{x}_i\}_{i=1}^{S}$ denote the states collected by the $S$-step student sampler, and let $\hat{\mathbf{x}}_0(\mathbf{x}_i,\sigma_i)$ denote the student's clean prediction from state $\mathbf{x}_i$ at noise level $\sigma_i$. For horizon $m$, $i_m$ denotes the selected student step and $\mathbf{z}_m$ the corresponding teacher completion. We use $\hat{\mathbf{x}}_{0,S}^{(\mathcal X)}=\mathcal{D}(\hat{\mathbf{x}}_0(\mathbf{x}_S,\sigma_S))$ and $\mathbf{z}_M^{(\mathcal X)}=\mathcal{D}(\mathbf{z}_M)$ for the final student and teacher outputs, where $\mathcal X$ denotes the decoded image representation. Given a differentiable reward model $R$, the terminal reward objective is $\mathcal{L}_\text{reward}(\theta_S)=-\mathbb{E}[R(\hat{\mathbf{x}}_{0,S}^{(\mathcal X)},c)]$, where $c$ is the conditioning signal.

\begin{algorithm}[t]
\caption{RATS: One Training Iteration}
\label{alg:training}
\begin{algorithmic}[1]
\Require Student $\theta_S$, teacher $\theta_T$, reward model $R$, decoder $\mathcal{D}$
\Require Horizons $\mathcal{H}{=}\{\bar{\sigma}_m\}_{m=1}^{M}$, weights $\{w_m^{(q)}\}_{m,q}$, shaping coefficient $\alpha$, EMA decay $\gamma$, gate temperature $\tau$
\State Sample noise $\mathbf{x}_1 \sim \mathcal{N}(\mathbf{0}, \mathbf{I})$ and prompt $c$
\State \textbf{Student rollout:} run the native $S$-step sampler; collect $\{\mathbf{x}_i,\hat{\mathbf{x}}_0(\mathbf{x}_i,\sigma_i)\}_{i=1}^{S}$ \Comment{with $\nabla$}
\State Decode student output: $\hat{\mathbf{x}}_{0,S}^{(\mathcal X)} = \mathcal{D}(\hat{\mathbf{x}}_0(\mathbf{x}_S,\sigma_S))$; compute $\mathcal{L}_\text{reward}$ and $R_S = R(\hat{\mathbf{x}}_{0,S}^{(\mathcal X)}, c)$
\For{$m = 1, \ldots, M$}
    \State Select student step $i_m$ and construct teacher completion $\mathbf{z}_m$ (Eq.~\ref{eq:sigma_match}) \Comment{no $\nabla$}
\EndFor
\State Decode teacher output: $\mathbf{z}_M^{(\mathcal X)} = \mathcal{D}(\mathbf{z}_M)$; compute $R_T = R(\mathbf{z}_M^{(\mathcal X)}, c)$ \Comment{no $\nabla$}
\State Compute the joint latent- and decoded-image shaping loss $\mathcal{L}_\text{shape}$ \Comment{Eq.~\ref{eq:shaping_loss}}
\State Compute reward gate $g(R_T,R_S)$ \Comment{Eq.~\ref{eq:gate}}
\State $\mathcal{L}_\text{total} \leftarrow \mathcal{L}_\text{reward} + \alpha\,g(R_T,R_S)\mathcal{L}_\text{shape}$ \Comment{Eq.~\ref{eq:total_loss}}
\State Backpropagate $\nabla_{\theta_S}\mathcal{L}_\text{total}$; update $\theta_S$ via optimizer
\State EMA update: $\theta_T \leftarrow \gamma \, \theta_T + (1 - \gamma) \, \theta_S$ \Comment{Eq.~\ref{eq:ema_update}}
\end{algorithmic}
\end{algorithm}

\subsection{Trajectory Shaping}
\label{sec:shaping}

This subsection describes how the teacher is constructed and how its state-conditioned endpoint targets jointly shape the student in latent and decoded image representations.

\noindent\textbf{EMA teacher and paired sampling.}
Let $\theta_S$ and $\theta_T$ denote the student and teacher parameters, respectively.
The teacher is initialized as a copy of the student, receives no gradient updates, and is kept in evaluation mode; its adapter parameters evolve only through EMA.
After each student update at iteration $k$, the teacher is updated via exponential moving average (EMA):
\begin{equation}
\label{eq:ema_update}
\theta_T^{(k+1)} \leftarrow \gamma \theta_T^{(k)} + (1-\gamma)\theta_S^{(k+1)}, \quad \gamma \in (0,1),
\end{equation}
where $\gamma$ is the decay coefficient. The EMA update is applied only to the adapter parameters, while the shared pretrained backbone remains fixed.
In each training iteration, the student follows its native $S$-step trajectory from initial noise $\mathbf{x}_1$. At selected states along this trajectory, the teacher uses a finer multi-step schedule to complete the stop-gradient student state toward the clean endpoint under the same prompt $c$.
This design provides a stable and up-to-date teacher reference while pairing each target with the state actually visited by the few-step student.

\noindent\textbf{Sigma-aligned multi-horizon matching.}
To provide intermediate trajectory supervision beyond a single terminal target, we construct teacher completion targets at multiple noise levels visited by the student.
A direct step-index alignment is inappropriate because the few-step student and the finer teacher completion use different schedules. We therefore index the supervision by the student's sigma coordinate rather than by sampler step.
Concretely, we define an ordered sequence of $M$ sigma horizons $\mathcal{H}=\{\bar{\sigma}_m\}_{m=1}^{M}$ spanning the denoising trajectory, with $\bar{\sigma}_1>\cdots>\bar{\sigma}_M$ from higher to lower noise.
For each horizon $\bar{\sigma}_m$, we select the nearest student state and ask the teacher to complete that state to the clean endpoint:
\begin{equation}
\label{eq:sigma_match}
i_m = \underset{i \in \{1,\dots,S\}}{\arg\min}\; \lvert \sigma_i - \bar{\sigma}_m \rvert, \qquad
\mathbf{z}_m = \operatorname{sg}\!\left[\mathcal{C}_T\!\left(\mathbf{x}_{i_m},\sigma_{i_m}\right)\right].
\end{equation}
Here, $\mathcal{C}_T(\mathbf{x},\sigma)$ denotes the EMA teacher completion from state $\mathbf{x}$ at noise level $\sigma$ to the clean endpoint under the shared condition $c$, and $\operatorname{sg}$ denotes stop-gradient. The teacher target and student prediction are aligned at the same starting sigma and, crucially, conditioned on the same visited state. Repeating this construction across $\mathcal H$ yields a schedule-agnostic multi-horizon correspondence without requiring teacher and student step indices to coincide.

The shaping objective is defined at the state, horizon, and representation levels as follows:
\begin{equation}
\label{eq:state_shaping}
\mathcal{L}_\text{state}(\hat{\mathbf{x}}_{0,m}^{(q)},\mathbf{z}_m^{(q)})
=\underbrace{\lambda_\text{cos}\!\left[1-\operatorname{cos}\!\left(\hat{\mathbf{x}}_{0,m}^{(q)},\mathbf{z}_m^{(q)}\right)\right]}_{\text{structural alignment}}
+\underbrace{\lambda_{\ell_2}\left\lVert\hat{\mathbf{x}}_{0,m}^{(q)}-\mathbf{z}_m^{(q)}\right\rVert_2^2}_{\text{magnitude alignment}}.
\end{equation}
\begin{equation}
\label{eq:space_shaping_components}
\mathcal{L}_{q}
=\sum_{m=1}^{M}w_m^{(q)}\,
\mathcal{L}_\text{state}(\hat{\mathbf{x}}_{0,m}^{(q)},\mathbf{z}_m^{(q)}),
\qquad q\in\{\mathcal Z,\mathcal X\}.
\end{equation}
\begin{equation}
\label{eq:shaping_loss}
\mathcal{L}_\text{shape}(\theta_S,\theta_T)
= \mathcal{L}_{\mathcal Z}
+ \lambda_{\mathcal X}\mathcal{L}_{\mathcal X}.
\end{equation}
For each horizon $m$ and representation $q\in\{\mathcal Z,\mathcal X\}$, $\hat{\mathbf{x}}_{0,m}^{(q)}$ and $\mathbf{z}_m^{(q)}$ denote the student clean prediction $\hat{\mathbf{x}}_0(\mathbf{x}_{i_m},\sigma_{i_m})$ and teacher completion $\mathbf{z}_m$ expressed in representation $q$, where $\mathcal Z$ and $\mathcal X$ denote the latent and decoded image representations, respectively. In Eq.~\ref{eq:state_shaping}, the cosine and squared $\ell_2$ terms preserve structural direction and state magnitude, respectively. Equation~\ref{eq:space_shaping_components} aggregates this discrepancy across horizons in each representation; $w_m^{(q)}$ emphasizes the active horizons, with $w_m^{(q)}=0$ for an inactive horizon and larger weights assigned to lower-noise predictions.

Equation~\ref{eq:shaping_loss} combines shaping in the latent and decoded image representations through $\mathcal{L}_{\mathcal Z}$ and $\mathcal{L}_{\mathcal X}$, with $\lambda_{\mathcal X}$ balancing their contributions. These training-time signals introduce no additional inference module, and all teacher targets are stop-gradient. When shaping is enabled, the student's gradient-tracking window covers at least the earliest selected step $i_1$, ensuring that gradients flow through all horizon-relevant denoising steps.

\subsection{Reward-Gated Modulation}
\label{sec:gate}

The trajectory shaping term offers informative supervision from the teacher's multi-step completions. However, applying it unconditionally risks rigid imitation, may undermine the student's self-correction capacity, and can reintroduce a teacher-bounded ceiling. We therefore modulate the complete shaping loss according to the relative reward performance between teacher and student.

Let $R(\cdot,c)$ be a frozen differentiable reward model that assigns a scalar preference score to a decoded output under condition $c$. The student and teacher scores are
\begin{equation}
\label{eq:reward_scores}
R_S=R\!\left(\hat{\mathbf{x}}_{0,S}^{(\mathcal X)},c\right),
\qquad
R_T=R\!\left(\mathbf{z}_M^{(\mathcal X)},c\right).
\end{equation}
The decoded student output $\hat{\mathbf{x}}_{0,S}^{(\mathcal X)}$ depends on $\theta_S$ through the differentiable $S$-step rollout from initial noise $\mathbf{x}_1$, whereas $\mathbf{z}_M^{(\mathcal X)}$ is treated as a stop-gradient teacher target.
The reward loss directly maximizes the expected preference score of the student output:
\begin{equation}
\label{eq:reward_loss}
\mathcal{L}_{\mathrm{reward}}(\theta_S)
=\mathbb{E}_{c,\mathbf{x}_1}\!\left[1-R_S\right]
=1-\mathbb{E}[R_S].
\end{equation}
The expectation is taken over training prompts and initial noise. This is the similarity-reward loss used in our implementation; it is equivalent to the generic reward-maximization form $-\mathbb{E}[R_S]$ up to an additive constant. Although the reward model and decoder are fixed, gradients propagate through both and through the differentiable student rollout to $\theta_S$.

We use the same terminal scores to determine how strongly the teacher should shape the current student trajectory. Specifically, we define the scalar reward gate
\begin{equation}
\label{eq:gate}
g(R_T,R_S)
=\frac{1}{1+\exp\!\left(-\operatorname{sg}[R_T-R_S]/\tau\right)},
\end{equation}
where $\tau>0$ controls the sharpness of the transition and $\operatorname{sg}$ denotes stop-gradient. Thus, $R_S$ remains differentiable in $\mathcal{L}_{\mathrm{reward}}$, while both scores are detached only when computing the gate. The resulting scalar jointly modulates all active realizations of $\mathcal{L}_\text{shape}$.

When the teacher reward exceeds the student reward by a margin relative to $\tau$, the gate approaches 1 and the shaping signal is emphasized, allowing the student to exploit the teacher's state-conditioned completions as a useful prior.
At equal rewards the gate is $1/2$; when the student exceeds the teacher by a sufficient margin, it decreases toward 0, suppressing the shaping term and preventing the student from being constrained by an inferior teacher trajectory.
In this way, teacher guidance is applied only when it is beneficial, yielding an adaptive balance between trajectory supervision and reward-driven improvement.
This conditional modulation is what allows RATS to benefit from trajectory-level supervision without sacrificing the student's ability to surpass the teacher.

\subsection{Training Objective}
\label{sec:objective}

At every training iteration, reward optimization and all shaping components are evaluated from the same student rollout and its teacher completions, and are optimized jointly:
\begin{equation}
\label{eq:total_loss}
\mathcal{L}_{\mathrm{total}}(\theta_S)
=
\mathcal{L}_{\mathrm{reward}}(\theta_S)
+
\alpha\, g(R_T, R_S)\, \mathcal{L}_{\mathrm{shape}}(\theta_S,\theta_T).
\end{equation}
where $\alpha > 0$ is the shaping coefficient that balances the two objectives.
Only the student's adapter parameters $\theta_S$ receive gradient updates; the teacher $\theta_T$ is updated solely via EMA (Eq.~\ref{eq:ema_update}).
The student receives the joint gradient $\nabla_{\theta_S}\mathcal{L}_\text{total}$ in a single update, after which the teacher is maintained through EMA. The full all-in-one training iteration is summarized in Algorithm~\ref{alg:training}.

\section{Experiments}
\label{experiments_results}

\subsection{Experimental Setup}
\label{sec:setup}

We adopt \textbf{FLUX1.0-dev} as the base model for image generation and \textbf{Wan2.1-T2V-1.3B-480P} as the base model for video generation. Both models are fine-tuned with LoRA~\cite{hu2022lora}, using HPSv2.1~\cite{wu2023human} as the reward model. For image generation, we use $\mathrm{CFG\ scale}=3.5$ and $\mathrm{shift}=3$, and report results under $3$, $5$, $8$, and $50$ NFEs. The generated images are evaluated using HPS, PickScore, and ImageReward.
For video generation, we adopt $\mathrm{CFG\ scale}=5$ and $\mathrm{shift}=3$, and report results under $5$ and $8$ NFEs. The generated videos are evaluated using VBench.
For both image and video generation, the EMA coefficient $\gamma$ is set to $0.999$, and the teacher inference steps are set to $50$. All experiments are conducted on the DanceGRPO dataset, which contains $50\mathrm{K}$ prompts, and training is performed on $8$ NVIDIA H100 GPUs.

\begin{table}[t]
\centering
\caption{Comparison of original and tuned models under different sampling steps. pink upward arrows indicate the improvement over FLUX1.0 dev at the same number of NFEs.}
\label{tab:step_comparison}
\begin{tabular}{lllll}
\toprule
Method & NFEs & HPS & PickScore & ImageReward \\
\midrule
Baseline & 3  & 18.43 & 19.99 & -0.3551 \\
\rowcolor{gray!10}Ours        & 3  & 32.15 \up{13.72} & 22.46 \up{2.47} & 1.0956 \up{1.4506} \\
\midrule
Baseline & 5  & 26.12 & 21.83 & 0.7443 \\
\rowcolor{gray!10}Ours        & 5  & 32.16 \up{6.04} & 22.68 \up{0.85} & 1.1337 \up{0.3894} \\
\midrule
Baseline & 8  & 28.43 & 22.42 & 0.9140 \\
\rowcolor{gray!10}Ours        & 8  & 33.81 \up{5.38} & 23.14 \up{0.72} & 1.3240 \up{0.4100} \\
\midrule
Baseline & 50 & 29.76 & 22.59 & 1.0037 \\
\rowcolor{gray!10}Ours        & 50 & 32.95 \up{3.19} & 22.79 \up{0.20} & 1.1544 \up{0.1507} \\
\bottomrule
\end{tabular}
\vspace{-1.5em}
\end{table}
\begin{table}[t]
\centering
\caption{Quantitative results for image generation. Pink upward arrows and Green downward arrows indicate the change of Ours relative to the best non-Ours method at the same number of NFEs.}
\resizebox{0.9\columnwidth}{!}{%
\begin{tabular}{lllll}
\toprule
Method & NFEs & HPS & PickScore & ImageReward \\
\midrule
Flux      & 3  & 18.43 & 19.99 & -0.3551 \\
Hyper-SD\cite{ren2024hyper}  & 3  & 28.80 & 22.14 &  0.9882 \\
SenseFlow\cite{ge2025senseflow} & 3  & 30.63 & 22.33 &  1.2030 \\
\rowcolor{gray!10}Ours      & 3  & 32.15 \up{1.52} & 22.46 \up{0.13} & 1.0956 \down{0.1074} \\
\midrule
Flux      & 5  & 26.12 & 21.83 &  0.7443 \\
Hyper-SD\cite{ren2024hyper}   & 5  & 27.83 & 22.08 &  1.0710 \\
SenseFlow\cite{ge2025senseflow}  & 5  & 30.99 & 22.53 &  1.2110 \\
\rowcolor{gray!10}Ours      & 5  & 32.16 \up{1.17} & 22.68 \up{0.15} & 1.3337 \up{0.1227} \\
\midrule
Flux      & 8  & 28.43 & 22.42 &  0.9140 \\
Hyper-SD\cite{ren2024hyper}   & 8  & 30.50 & 22.76 &  1.0410 \\
SenseFlow\cite{ge2025senseflow}  & 8  & 30.99 & 22.59 &  1.1720 \\
\rowcolor{gray!10}Ours      & 8  & 33.81 \up{2.82} & 23.14 \up{0.38} & 1.3240 \up{0.1520} \\
\midrule
Flux      & 50 & 29.76 & 22.59 &  1.0037 \\
Hyper-SD\cite{ren2024hyper}  & 50 & 30.01 & 22.51 &  0.9461 \\
SenseFlow\cite{ge2025senseflow}  & 50 & 30.69 & 22.31 &  1.0810 \\
\rowcolor{gray!10}Ours      & 50 & 32.95 \up{2.26} & 22.79 \up{0.26} & 1.1544 \up{0.0734} \\
\bottomrule
\end{tabular}%
}
\label{imageGen-comparsion}
\end{table}
\begin{figure*}[t]
    \centering
    \includegraphics[width=0.95\linewidth]{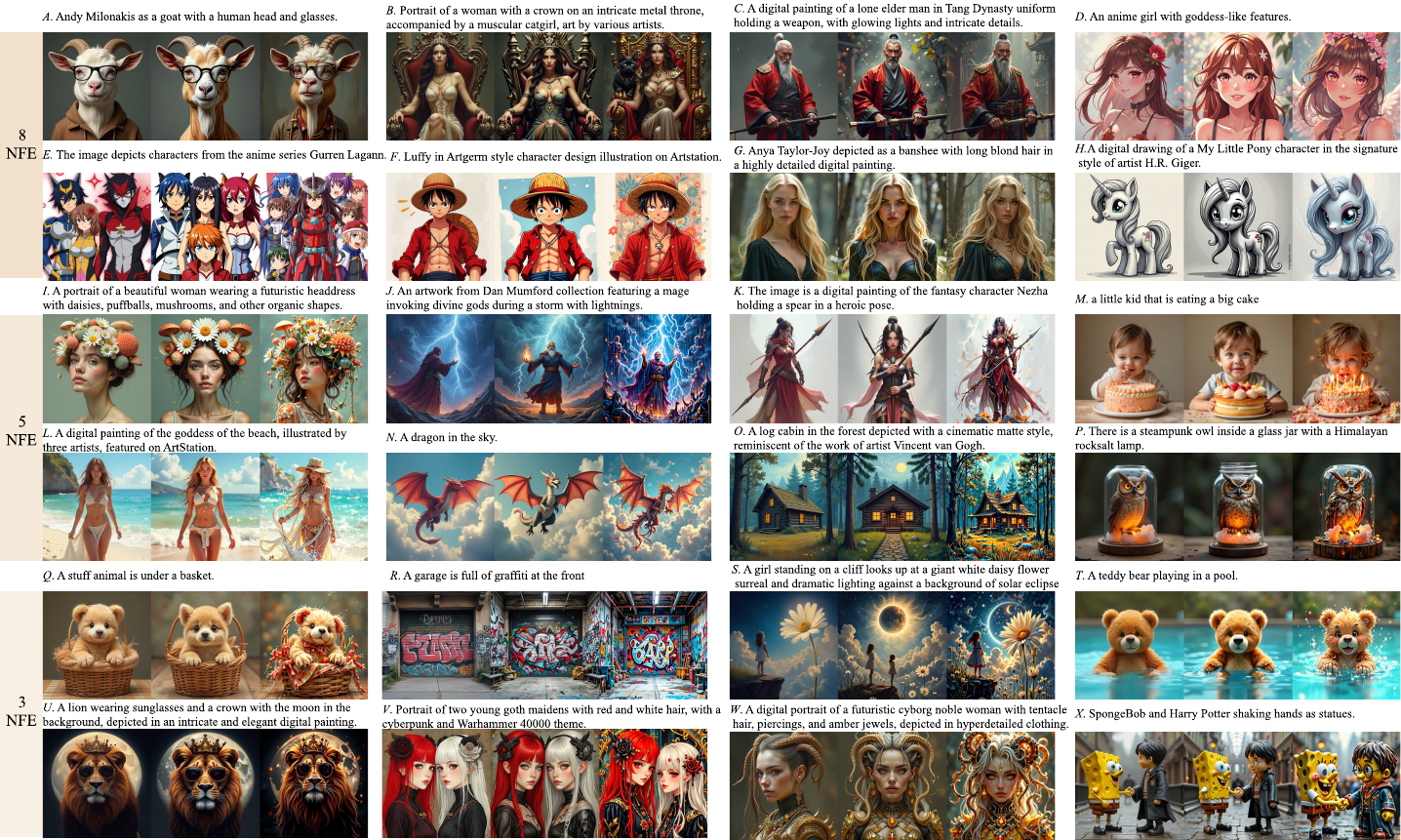}
    \caption{Qualitative results on Flux. Few-step image generation: left—Hyper-Flux, middle—SenseFlow, \textcolor{red}{right—Ours}.}
    \label{fig:qualititive_exp}
    \Description{3}
\end{figure*}

\begin{figure*}[t]
    \centering
    \includegraphics[width=0.95\linewidth]{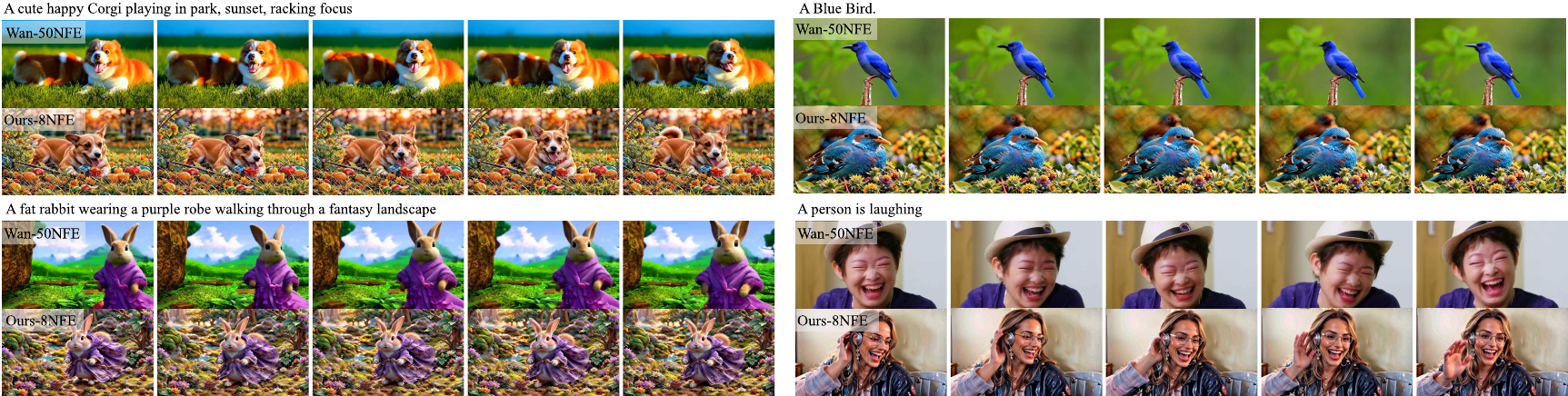}
    \caption{Qualitative comparison between our method with 8 NFEs and Wan with 50 NFEs. Our method produces consistently better visual quality, stronger text alignment, and more coherent motion dynamics than Wan-50NFE.}
    \label{fig:wan_case}
    \Description{4}
\end{figure*}
\subsection{Quantitative Results for Few-Step Image Generation}

\subsubsection{Comparison with the Baseline Model}
\begin{table}[t]
\centering
\caption{Overall evaluation results. Pink upward arrows indicate the improvement over Wan at the same number of NFEs.}
\begin{tabular}{lllll}
\toprule
Method & NFEs & Quality Score & Semantic Score & Total Score \\
\midrule
Wan  & 50 & 83.08 & 62.93 & 79.05 \\
\rowcolor{gray!10}Ours & 50 & 83.99 \up{0.91} & 65.99 \up{3.06} & 80.40 \up{1.35} \\
\midrule
Wan  & 8  & 77.82 & 48.74 & 72.01 \\
\rowcolor{gray!10}Ours & 8  & 82.66 \up{4.84} & 70.35 \up{21.61} & 80.20 \up{8.19} \\
\midrule
Wan  & 5  & 73.64 & 34.10 & 65.73 \\
\rowcolor{gray!10}Ours & 5  & 81.23 \up{7.59} & 67.78 \up{33.68} & 78.53 \up{12.80} \\
\bottomrule
\end{tabular}
\label{vbench-overall}
\vspace{-1em}
\end{table}
\begin{table}[t]
\centering
\caption{Quantitative results for video generation. Blue upward arrows and red downward arrows indicate the change of Ours relative to Wan at the same number of NFEs.}
\vspace{-1em}
\resizebox{\linewidth}{!}{%
\begin{tabular}{llcccccccccc}
\toprule
Method & NFEs
& AQ & Color & HA & IQ
& MO & OC & OCons & Scene & SR & SC \\
\midrule
Wan & 5
& 41.42 & 61.65 & 34.00 & 39.30
& 7.85 & 21.75 & 13.32 & 5.45 & 22.57 & 90.75 \\

\rowcolor{gray!10}Ours & 5
& \scoreup{71.81}{30.39} & \scoreup{90.26}{28.61}
& \scoreup{67.00}{33.00} & \scoreup{72.70}{33.40}
& \scoreup{67.75}{59.90} & \scoreup{75.47}{53.72}
& \scoreup{22.15}{8.83} & \scoreup{26.16}{20.71}
& \scoreup{83.50}{60.93} & \scoreup{98.07}{7.32} \\

\midrule

Wan & 8
& 50.31 & 81.47 & 61.00 & 51.15
& 23.01 & 41.45 & 17.15 & 7.84 & 47.16 & 91.73 \\

\rowcolor{gray!10}Ours & 8
& \scoreup{74.78}{24.47} & \scoreup{87.83}{6.36}
& \scoreup{72.00}{11.00} & \scoreup{75.40}{24.25}
& \scoreup{73.90}{50.89} & \scoreup{84.57}{43.12}
& \scoreup{23.25}{6.10} & \scoreup{24.37}{16.53}
& \scoreup{84.97}{37.81} & \scoreup{98.32}{6.59} \\

\midrule

Wan & 50
& 58.66 & 84.59 & 72.00 & 66.02
& 54.87 & 68.51 & 22.85 & 17.65 & 67.13 & 93.90 \\

\rowcolor{gray!10}Ours & 50
& \scoreup{66.89}{8.23} & \scoredown{74.52}{10.07}
& \scoreup{75.00}{3.00} & \scoreup{76.32}{10.30}
& \scoreup{70.35}{15.48} & \scoreup{78.56}{10.05}
& \scoreup{23.85}{1.00} & \scoreup{22.02}{4.37}
& \scoreup{67.40}{0.27} & \scoreup{96.27}{2.37} \\
\bottomrule
\end{tabular}%
}
\label{video_detail}
\vspace{-1.0em}
\end{table}
Table~\ref{tab:step_comparison} shows consistent gains over FLUX1.0-dev at 3, 5, 8, and 50 NFEs, with the largest gains at 3 NFEs: 13.72 in HPS, 2.47 in PickScore, and 1.4506 in ImageReward. Gains remain clear at 50 NFEs, indicating improved few-step sampling efficiency without compromising standard multi-step quality.

\subsubsection{Comparison with Strong Few-Step Baselines}

Table~\ref{imageGen-comparsion} further compares our method with strong few-step baselines, including Hyper-SD and SenseFlow. Our model achieves the best HPS and PickScore across all evaluated step settings, and also attains the best ImageReward at 5, 8, and 50 NFEs. Although SenseFlow obtains a slightly higher ImageReward at 3 NFEs, our method still delivers the strongest overall performance in this highly constrained setting.
These results show that our approach is not only effective relative to the original FLUX1.0-dev baseline, but also highly competitive against state-of-the-art few-step generation methods. More importantly, our advantage is not restricted to one specific metric or one specific sampling budget: it consistently transfers across both reward-aligned and external evaluation metrics, while remaining robust under larger-step inference. Since HPSv2.1 is used as the reward model during training, HPS can be regarded as an in-domain metric, while PickScore and ImageReward serve as out-of-domain metrics for evaluating generalization beyond the training reward. 
In particular, \textbf{under 5-step and 8-step inference, our method achieves performance comparable to or even better than the 50-step FLUX baseline on multiple metrics}.
We attribute this behavior to the reward-aware nature of our design. Rather than forcing the student to mimic the teacher uniformly, we introduce a reward-aware gating mechanism that adaptively balances distillation learning and preference-oriented optimization. Once the student has acquired a certain level of few-step generation capability, improving visual quality and alignment with human preferences becomes more important than overfitting to the teacher’s trajectory itself.
\begin{figure*}[t]
    \centering
    \includegraphics[width=0.9\linewidth]{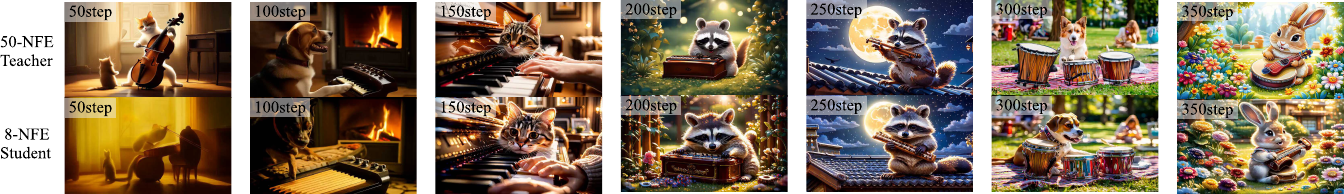}
    \caption{Teacher--student quality comparison on the first frames of generated videos throughout training. Our method progressively enables the few-step student to outperform the multi-step teacher in generation quality.}
    \label{fig:Teacher-Student}
    \Description{5}
\end{figure*}
\subsection{Few-Step Video Generation}
\begin{table}[t]
\centering
\caption{Ablation study on few-step image generation under different sampling budgets. Results are reported at 3, 5, and 8 NFEs, evaluated by HPS, PickScore, and ImageReward.}
\label{tab:ablation}
\vspace{-1em}
\resizebox{0.9\linewidth}{!}{
\begin{tabular}{ccccc}
\toprule
NFEs & Method & HPS & PickScore & ImageReward \\
\midrule
3 & Reward-Only        & 20.33 & 20.26 & 0.1353 \\
3 & Distillation-Only  & 24.34 & 20.61 & 0.4523 \\
3 & Reward+Distillation(no gate)        &26.72  & 21.35 & 0.9981 \\
\rowcolor{gray!10} 3 & Ours               & 32.15 & 22.46 & 1.0956 \\
\midrule
5 & Reward-Only        & 26.88 & 22.01 & 0.8912 \\
5 & Distillation-Only  & 27.01 & 22.38 & 0.8013 \\
5 & Reward+Distillation(no gate)        &29.97  & 22.35 & 1.0718 \\
\rowcolor{gray!10} 5 & Ours               & 32.16 & 22.68 & 1.1337 \\
\midrule
8 & Reward-Only        & 29.86 & 22.77 & 0.9560 \\
8 & Distillation-Only  & 30.10 & 22.04 & 0.9230 \\
8 & Reward+Distillation(no gate)       &33.18  & 22.82 & 1.1568 \\
\rowcolor{gray!10} 8 & RATS           & 33.81 & 23.14 & 1.3240 \\
\bottomrule
\end{tabular}
}
\end{table}
\begin{table}[t]
\centering
\caption{Ablation results about $\alpha$ values.}
\label{tab:alpha_results}
\vspace{-1em}
\resizebox{0.70\columnwidth}{!}{%
\begin{tabular}{lcccc}
\hline
NFEs & $\alpha$ & HPS & PickScore & ImageReward \\
\hline
\multirow{4}{*}{3}
& 0.2 & 32.33 & 22.08 & 1.0082 \\
& 0.5 & 32.08 & 22.47 & 1.1052 \\
& 1   & 31.88 & 22.06 & 1.0826 \\
& \cellcolor{gray!10}2 & \cellcolor{gray!10}32.15 & \cellcolor{gray!10}22.46 & \cellcolor{gray!10}1.0956 \\
\hline
\multirow{4}{*}{5}
& 0.2 & 31.78 & 22.09 & 1.1615 \\
& 0.5 & 32.09 & 22.51 & 1.2058 \\
& \cellcolor{gray!10}1 & \cellcolor{gray!10}32.16 & \cellcolor{gray!10}22.68 & \cellcolor{gray!10}1.3337 \\
& 2   & 31.87 & 22.36 & 1.0891 \\
\hline
\multirow{4}{*}{8}
& 0.2 & 33.07 & 22.56 & 1.1730 \\
& 0.5 & 32.24 & 22.67 & 1.1829 \\
& \cellcolor{gray!10}1 & \cellcolor{gray!10}33.81 & \cellcolor{gray!10}23.14 & \cellcolor{gray!10}1.3240 \\
& 2   & 32.13 & 22.97 & 1.1622 \\
\hline
\end{tabular}
}
\vspace{-1em}
\end{table}

\begin{table}[t]
\centering
\caption{Efficiency comparison of different methods. \textcolor{tabpink}{Best} and \textcolor{tabblue}{second-best} results are highlighted.}
\label{tab:efficiency}
\vspace{-1em}
\resizebox{\columnwidth}{!}{%
\begin{tabular}{lccccccc}
\toprule
Method & \makecell{Peak Memory\\(GB)} & \makecell{Total Steps\\(K)} & \makecell{Total Time\\(h)} & \makecell{Total FLOPs\\(PFLOPs)} & \makecell{Extra\\Data} & \makecell{Few-Step\\Generation} & \makecell{Preference\\Align.} \\
\midrule
SenseFlow 
& 78.87 
& 12.0 
& 24.35 
& 9615
& Yes 
& Yes 
& No  \\

DanceGRPO 
& \textcolor{tabpink}{34.05} 
& \textcolor{tabpink}{0.2}  
& \textcolor{tabblue}{11.78} 
& \textcolor{tabpink}{321}
& No  
& No  
& Yes \\

\rowcolor{gray!10}
Ours      
& \textcolor{tabblue}{45.26}
& \textcolor{tabpink}{0.2}
& \textcolor{tabpink}{1.52}
& \textcolor{tabblue}{3926}
& \textcolor{tabpink}{No}  
& \textcolor{tabpink}{Yes} 
& \textcolor{tabpink}{Yes} \\
\bottomrule
\end{tabular}%
}
\vspace{-0.5em}
\end{table}

\subsubsection{Overall Comparison}
As shown in Table~\ref{vbench-overall}, our method consistently improves Wan across all evaluated sampling budgets, including 5, 8, and 50 NFEs. Similar to the image generation results, the gains are most significant in the low-step regime. At 5 NFEs, our method improves the Quality Score, Semantic Score, and Total Score by 7.59, 33.68, and 12.80 points, respectively. At 8 NFEs, the corresponding gains remain large, reaching 4.84, 21.61, and 8.19 points. Even at 50 NFEs, our method still brings consistent improvements of 0.91, 3.06, and 1.35 points. These results indicate that our method is particularly effective for few-step video generation, while also preserving gains under standard multi-step inference.

More importantly, \textbf{our few-step model can already match or even surpass the strong multi-step baseline}. With only 8 NFEs, our method achieves a Total Score of 80.20, outperforming Wan at 50 NFEs, which obtains 79.05. Even under 5-NFEs inference, our method reaches 78.53 in Total Score, which is already very close to the 50-step Wan baseline, while achieving a substantially higher Semantic Score (67.78 vs. 62.93). This suggests that the proposed method significantly improves generation efficiency, especially in terms of semantic alignment under constrained sampling budgets.

\subsubsection{Detailed VBench Analysis}
The detailed VBench results in Table~\ref{video_detail} further support this conclusion. Under both 5-step and 8-step inference, our method improves a broad range of dimensions, including Aesthetic Quality, Background Consistency, Color, Human Action, Imaging Quality, Multiple Objects, Object Class, Overall Consistency, Scene, Spatial Relationship, Subject Consistency, and Temporal Style. In particular, the large gains on Multiple Objects, Object Class, Spatial Relationship, and Human Action indicate that our method substantially enhances semantic controllability and compositional understanding in few-step video generation. Although a few dimensions, such as Appearance Style and Dynamic Degree, remain slightly lower, the improvements on most quality- and semantic-related dimensions are much larger, leading to clear gains in the overall evaluation.

\subsection{Ablation}

\subsubsection{Effectiveness of reward-aware training.}
Table~\ref{tab:ablation} evaluates the contribution of each component under different sampling budgets. Both \textit{Reward-Only} and \textit{Distillation-Only} improve performance to some extent, but neither of them alone matches the full model. Simply combining the two objectives without gating further improves the results, yet still remains consistently below our final method. Importantly, our method not only performs best on HPS, which is closely related to the training reward, but also consistently achieves the highest PickScore and ImageReward, showing that the gain is not merely due to overfitting to the reward model, but instead reflects genuine improvements in generation quality. As the number of sampling steps increases, the margin between the full model and the ablated variants gradually decreases, while remaining consistently positive. This trend suggests that the proposed design is particularly beneficial for few-step generation, where efficient guidance and preference alignment are both crucial.

\subsubsection{Ablation on $\alpha$.}
Table~\ref{tab:alpha_results} studies the effect of the weighting coefficient $\alpha$. We observe that the optimal value depends on the sampling budget, but intermediate values generally work better than overly small or overly large ones. Overall, the results show that our method is robust to a reasonable range of $\alpha$, while moderate values provide the most balanced performance across different metrics.

\subsection{Qualitative Evaluation}

\subsubsection{Qualitative Result for Image Generation}
As shown in Figure~\ref{fig:qualititive_exp}, compared with Hyper-SD and SenseFlow, our method (the \textbf{third column} in each group) consistently generates images with more complete structures, cleaner local details, and more faithful semantic alignment under few-step inference. Across diverse prompts involving portraits, stylized characters, complex costumes, fantasy scenes, and compositional backgrounds, our results are generally sharper and more balanced in both global composition and local fidelity. In particular, our method better preserves visually important regions such as facial identity, clothing texture, object boundaries, and background richness, indicating that the proposed reward-aware trajectory shaping improves not only sampling efficiency but also human-preferred visual quality in challenging low-NFE regimes.

\subsubsection{Qualitative Result for Video Generation}
The qualitative video results show an equally clear advantage and further demonstrate the generality of our method across image and video generation. As shown in Figure~\ref{fig:wan_case}, our 8-NFE model already produces perceptually better videos than Wan with 50 inference steps, with clearer subjects, richer backgrounds, and stronger semantic consistency across the three examples. In the corgi case, our result yields a more faithful dog appearance with sharper fur texture and a more vivid park scene; in the fantasy rabbit case, it preserves a clearer subject identity and more detailed environmental rendering; in the laughing-person case, it better captures facial expression and portrait clarity. More importantly, Figure~\ref{fig:Teacher-Student} shows that, as training proceeds, our few-step student progressively \textbf{surpasses the multi-step teacher in generation quality}.

\subsection{Efficiency analysis of different methods}

As shown in Table~\ref{tab:efficiency}, SenseFlow requires 12K steps, 24.35 hours, 9615 PFLOPs, and extra data. DanceGRPO reduces total compute to 321 PFLOPs but takes 11.78 hours and does not provide few-step acceleration. In contrast, RATS completes its 0.2K-step training in 1.52 hours with 3926 PFLOPs and no extra data, while supporting both few-step generation and preference alignment.

\subsection{Student Surpasses Teacher in RATS}

\begin{figure}[ht]
    \centering
    \includegraphics[width=0.95\linewidth]{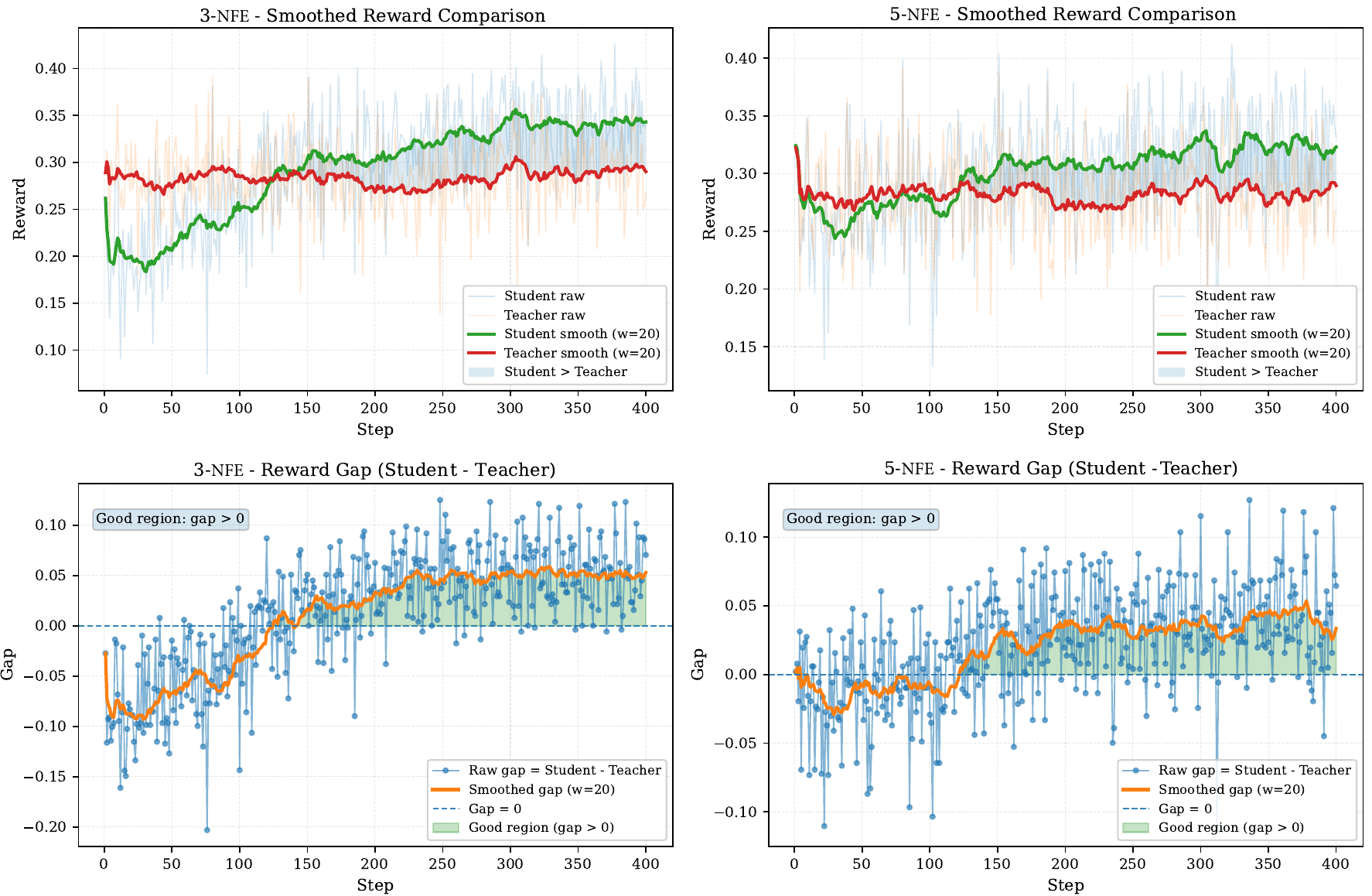}
    \caption{Smoothed reward comparison (top) and reward gap (bottom) for 3-step and 5-step settings. 
The student model gradually surpasses the teacher, as shown by higher smoothed rewards in the later stages.}
    \label{fig:student_teacher_reward_and_gap_3_5_8step_horizontal_marked}
    \Description{6}
\end{figure}

Figure~\ref{fig:student_teacher_reward_and_gap_3_5_8step_horizontal_marked} shows that the student progressively surpasses the teacher during training, as evidenced by higher smoothed rewards and a consistently positive reward gap. The effect is particularly pronounced in the few-step regime, demonstrating the effectiveness of our framework under constrained sampling budgets.

\section{Conclusion}
In this paper, we presented \textbf{RATS}, a reward-aware trajectory shaping framework for few-step generation. Instead of relying only on final-output reward or rigid teacher imitation, RATS introduces sigma-aligned multi-horizon trajectory matching together with a reward-aware gate, enabling the student to benefit from teacher guidance when useful while still surpassing the teacher when reward optimization becomes more informative. Extensive experiments on both image and video generation show that RATS consistently improves low-NFE generation, with especially clear gains in the most challenging few-step regimes. These results suggest that effective few-step alignment depends not only on optimizing the final reward, but also on shaping the intermediate denoising trajectory during training. We hope this work provides a useful direction for bridging fast inference and high-quality aligned generation.


\bibliographystyle{ACM-Reference-Format}
\bibliography{main}

\appendix
\end{document}